\documentclass[journal]{IEEEtran}
\usepackage{stfloats} 
\usepackage{color}
\usepackage{tabularx}
\usepackage{booktabs}
\usepackage{graphicx} 
\usepackage{hyperref}
\usepackage{amsmath}
\usepackage{amssymb}
\usepackage{dsfont}
\usepackage{float} 
\usepackage{subfig} 
\usepackage{algorithm}
\usepackage{algorithmic}
\usepackage{multirow}
\usepackage{graphicx, subcaption}

\begin{document}
\title{Label-Consistent Dataset Distillation with Detector-Guided Refinement}
\author{Yawen Zou$^{1}$, Guang Li$^{2}$, Zi Wang$^{3}$, Chunzhi Gu$^{4}$, Chao Zhang$^{1}$ \\
$^{1}$University of Toyama, 
$^{2}$Hokkaido University, \\
$^{3}$Niigata University,
$^{4}$Toyohashi University of Technology
}
\maketitle

\begin{abstract}
Dataset distillation (DD) aims to generate a compact yet informative dataset that achieves performance comparable to the original dataset, thereby reducing demands on storage and computational resources. Although diffusion models have made significant progress in dataset distillation, the generated surrogate datasets often contain samples with label inconsistencies or insufficient structural detail, leading to suboptimal downstream performance. To address these issues, we propose a detector-guided dataset distillation framework that explicitly leverages a pre-trained detector to identify and refine anomalous synthetic samples, thereby ensuring label consistency and improving image quality. Specifically, a detector model trained on the original dataset is employed to identify anomalous images exhibiting label mismatches or low classification confidence. For each defective image, multiple candidates are generated using a pre-trained diffusion model conditioned on the corresponding image prototype and label. The optimal candidate is then selected by jointly considering the detector’s confidence score and dissimilarity to existing qualified synthetic samples, thereby ensuring both label accuracy and intra-class diversity. Experimental results demonstrate that our method can synthesize high-quality representative images with richer details, achieving state-of-the-art performance on the validation set.
\end{abstract} 

Large-scale datasets have driven significant advances in artificial intelligence (AI). However, they also pose significant problems, including increased storage demands, high transmission burdens, and considerable computational costs. To mitigate these limitations, extensive research has explored data-efficient learning techniques. Among them, dataset distillation has garnered significant attention for its ability to synthesize compact and representative surrogate datasets while maintaining comparable training efficacy to the original large-scale datasets \cite{wang2018dataset,li2022awesome,li2025ranking}. By substantially reducing data volume, dataset distillation enables more efficient model training and improves scalability, particularly in resource-constrained settings. Moreover, it has demonstrated promising applications across diverse domains, including continual learning \cite{gu2024ssd,yang2023efficient}, privacy preserving \cite{vinaroz2023dpkip,zheng2024differentially}, and medical data \cite{li2024medical,yu2024progressive}.

\begin{figure}[t] 
\centering 
\includegraphics[width=0.48\textwidth]{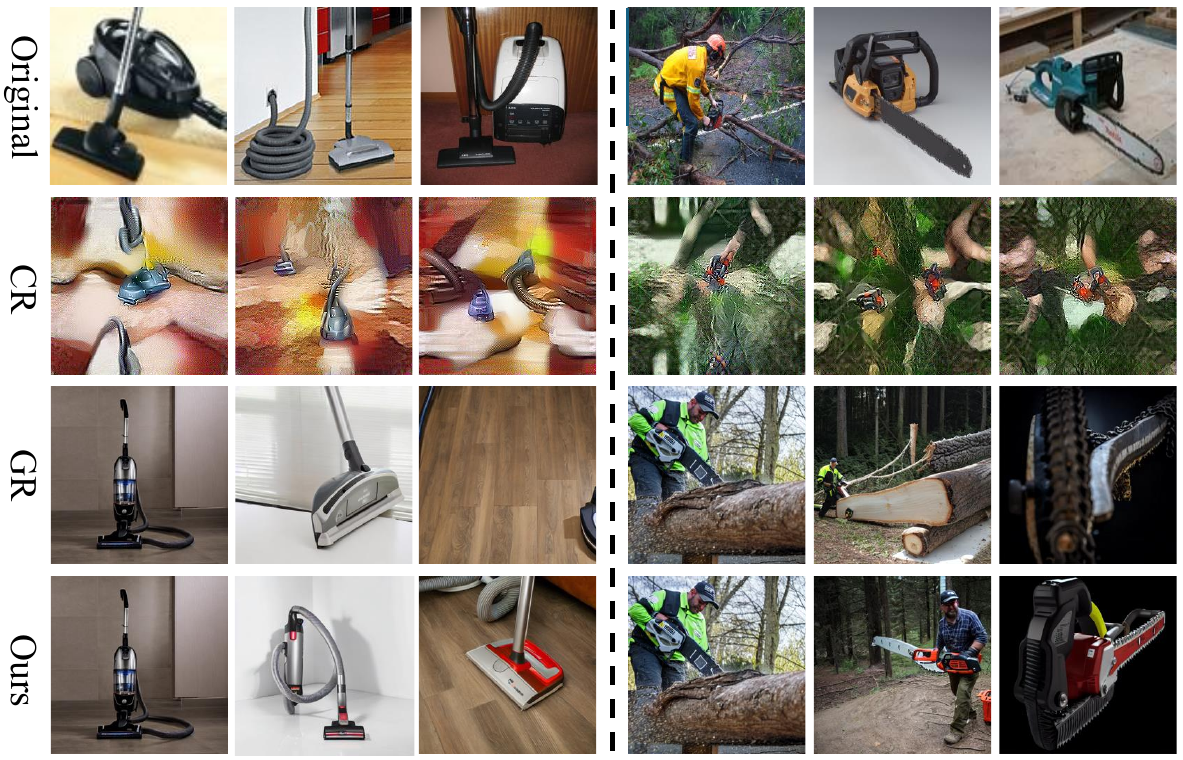} 
\caption{Visualization comparison of different dataset distillation methods. 
CR shows distilled images obtained by the conventional method SRe$^{2}$L~\cite{yin2023sre2l}, 
while GR denotes images synthesized by the diffusion-based generative method D$^{4}$M~\cite{su2024d}.
} 
\label{fig1} 
\end{figure}

Previous works are mainly designed to improve dataset distillation through meta-learning strategies \cite{nguyen2021kip,zhou2022dataset} and matching-based techniques \cite{zhao2021datasetcondensation,liu2024att,zhao2023distribution,wang2022cafe}. However, a fundamental limitation of these approaches lies in their poor generalization across different network architectures and their inefficiency when applied to high-resolution images (e.g., $256 \times 256$ pixels). These limitations are largely attributed to the inherent difficulty of directly optimizing in the high-dimensional pixel space, which often results in overfitting to the specific architecture used during distillation \cite{cazenavette2023glad} and significantly compromises scalability concerning increasing image resolution and dataset size \cite{lei2023comprehensive}. Consequently, conventional dataset distillation frameworks typically incur substantial computational overhead and struggle to scale to large-scale datasets such as ImageNet-1K. To address these problems, recent generative approaches, including Generative Adversarial Networks (GANs) \cite{cazenavette2023glad,zhong2024hierarchical} and diffusion models \cite{su2024d,gu2024efficient}, have been proposed to synthesize compact datasets by optimizing in the latent space rather than the pixel space. By decoupling the generation process from specific network architectures, these generative methods effectively alleviate the architectural dependency inherent in traditional dataset distillation and enable the synthesis of realistic, high-resolution images with reduced computational overhead \cite{su2024d,moser2024ld3m}. 

Despite these advantages, generative DD methods still face notable challenges, including label noise and insufficient class-discriminative details, which hinder downstream task performance. As illustrated in Fig. \ref{fig1}, existing methods like $\text{D}^4\text{M}$ \cite{su2024d} often generate images that lack sufficient structural details or fail to include the intended target objects, making it difficult for models to extract meaningful features. For example, samples generated for the vacuum cleaner and chainsaw classes frequently contain only background textures or incomplete object structures, resulting in limited class discriminability. Such anomalous images are often associated with incorrect labels or low confidence scores. Quantitatively, for images synthesized by $\text{D}^4\text{M}$ \cite{su2024d} with an images per class (IPC) setting of 10, up to 12\% of the labels are incorrect, and 5\% of the samples exhibit confidence scores below 0.7. Such low-quality images can significantly impair model performance and reliability in downstream tasks.

To address these issues, we introduce an anomaly detection framework to identify and refine defective synthetic samples, thereby enhancing the overall quality of the distilled dataset. Unlike conventional approaches that overlook the quality control of generated images, our method leverages a pre-trained detector to guide the generative process, enabling the synthesis of datasets with enhanced structural coherence and high label consistency. Specifically, we first extract representative prototypes from the original dataset and employ a Latent Diffusion Model (LDM) to generate a surrogate dataset. Subsequently, a detector model, trained on the original dataset, is utilized to identify anomalous samples within the synthetic dataset. To refine these defective images, we generate multiple candidate images for each anomalous sample and select the top-$k$ candidates based on the detector's confidence scores. From these, the image exhibiting the greatest dissimilarity to existing qualified samples is selected, thereby ensuring dataset representativeness while improving intra-class diversity.

The key contributions of this work are summarized as follows:
\begin{itemize}
\item We introduce a detector-guided dataset distillation framework that integrates a pre-trained detector to identify and refine defective synthetic samples, effectively mitigating label noise and structural inconsistencies commonly observed in generative distillation.

\item We propose a targeted refinement strategy that generates multiple image variations for each anomalous sample and selects the most dissimilar candidate relative to existing qualified samples, thereby enhancing intra-class diversity and representativeness of the distilled dataset.

\item Extensive experiments demonstrate that our method significantly improves the quality of synthetic datasets and achieves superior performance in downstream classification tasks compared to existing baselines.
\end{itemize}

\begin{figure*}[t] 
\centering 
\includegraphics[width=0.7\textwidth]{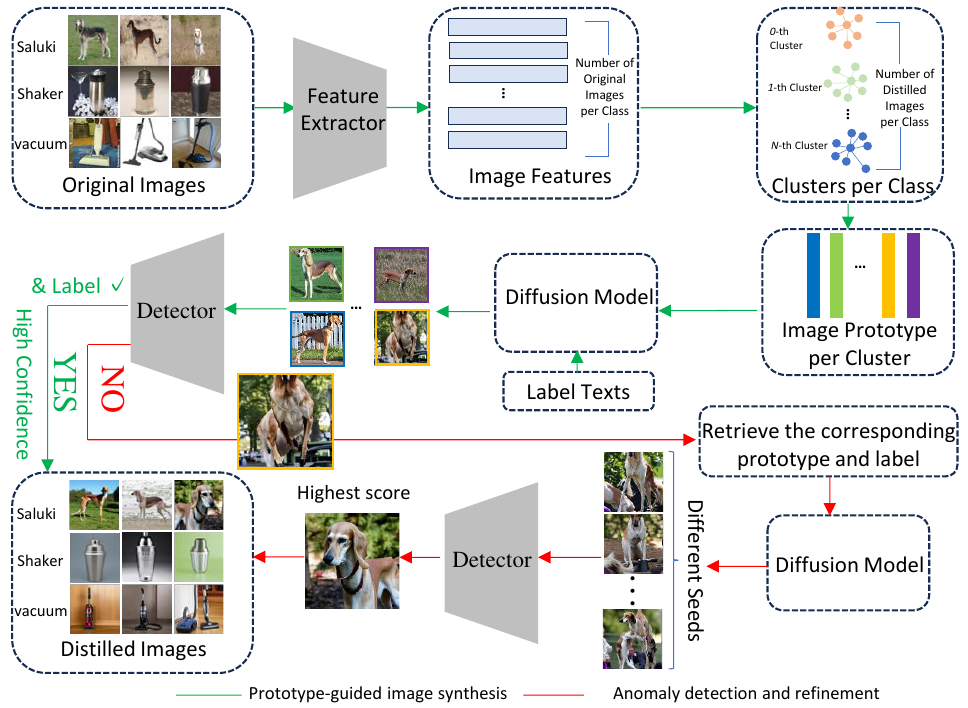} 
\caption{Overview of the proposed framework. It comprises two primary modules: prototype-guided image synthesis and anomaly detection with iterative refinement. The first module synthesizes images guided by class-specific prototypes using a diffusion model. In contrast, the second module identifies defective samples, which are refined through candidate generation and selection by jointly considering the detector’s confidence and the dissimilarity to existing qualified samples.
} 
\label{fig2} 
\end{figure*}

\section{Related Work}
Dataset distillation aims to condense a large-scale dataset into a small yet informative surrogate that achieves performance comparable to the original data, thereby enabling faster training and improved scalability, particularly in resource-constrained settings. The concept of dataset distillation was introduced by Wang et al. \cite{wang2018dataset}, and has since been widely adopted and extended by subsequent methods, including meta-learning frameworks \cite{nguyen2021kip,zhou2022dataset}, matching-based techniques \cite{zhao2021datasetcondensation,liu2024att,zhao2023distribution,wang2022cafe}, and more recent generative methods \cite{cazenavette2023glad,zhong2024hierarchical,su2024d,gu2024efficient}.

\textbf{Meta-learning approaches} employ a bi-level optimization framework, where the outer loop minimizes the performance loss on the real dataset, while the inner loop concurrently updates the model using the synthetic data \cite{nguyen2021kip,zhou2022dataset}. These methods can be categorized into two groups based on the inner-loop optimization strategy: backpropagation through time (BPTT) and kernel ridge regression (KRR). Wang et al. \cite{wang2018dataset} first proposed formulating distilled images as learnable parameters within a meta-learning framework, enabling the optimization of synthetic datasets for downstream tasks. To alleviate gradient instability and high computational costs, Feng et al. \cite{feng2023embarassingly} proposed Random Truncated BPTT (RaT-BPTT), which integrates gradient truncation with randomization to enhance stability and efficiency, particularly when addressing long-term dependencies.
To further alleviate the computational burden, Nguyen et al. \cite{nguyen2020dataset} replaced the neural network in the inner loop with a kernel model, eliminating the need for explicit meta-gradient backpropagation. Building upon this, they incorporated a kernel-based meta-learning framework into infinitely wide convolutional networks to enhance dataset distillation performance \cite{nguyen2021dataset}.

\textbf{Matching-based methods} optimize distilled datasets through two primary strategies: parameter matching and distribution matching. Parameter matching aligns model parameters learned from both the original and synthetic datasets \cite{zhao2021datasetcondensation,liu2024att}. 
Zhao et al. \cite{zhao2020dataset} proposed matching the training gradients induced by the synthetic and real datasets to achieve comparable performance. 
In contrast, distribution matching directly aligns the distributions of learned feature representations extracted from the real and synthetic datasets \cite{zhao2023distribution,wang2022cafe}. Zhao et al. \cite{zhao2023dataset} employed Maximum Mean Discrepancy (MMD) as the distance metric to align the class-wise feature distributions between synthetic and real datasets. Wang et al. \cite{wang2022cafe} further proposed CAFE, which aligns multi-layer feature statistics to achieve more comprehensive semantic consistency.
However, these methods still face scalability challenges on large-scale datasets like ImageNet-1K and often exhibit poor generalization across diverse network architectures.

\textbf{Generative Methods }leverage GANs and diffusion models to synthesize surrogate datasets. GAN-based approaches optimize synthetic images directly in the pixel space, whereas diffusion-based methods typically operate in the latent space, enabling more efficient training and improved distillation performance.
By synthesizing data in latent space, diffusion-based methods can produce more representative surrogate datasets while significantly reducing computational costs.
Gu et al. \cite{gu2024efficient} introduced additional minimax criteria to train the Diffusion Transformer (DiT) model for synthesizing a high-quality surrogate dataset. Zou et al. \cite{zou2025vlcp} integrated class-specific textual and visual prototypes to condition the diffusion model, enabling it to synthesize representative and diverse images and thereby achieve superior performance. Su et al. \cite{su2024d} proposed a prototype-guided approach, where K-means clustering is applied to extract class prototypes, which are subsequently combined with label text prompts to guide image synthesis via the Latent Diffusion Model (LDM).

\section{Methodology}
\subsection{Preliminary}
Dataset distillation aims to synthesize a small yet informative dataset 
\(\mathcal{S} = \left\{ (\tilde{x}_j, \tilde{y}_j) \right\}_{j=1}^{N_\mathcal{S}}\) 
that preserves the essential information of a large-scale dataset 
\(\mathcal{T} = \left\{ (x_i, y_i) \right\}_{i=1}^{N_\mathcal{T}}\), 
where \(\tilde{x}_j\) and \(x_i\) represent images, \(\tilde{y}_j\) and \(y_i\) denote their corresponding class labels, 
and \(N_\mathcal{S} \ll N_\mathcal{T}\), following \cite{zhao2020dataset,wang2018dataset}. 
The primary objective of dataset distillation is to enable models trained on the distilled dataset \(\mathcal{S}\) to achieve similar test performance to those trained on the full dataset \(\mathcal{T}\), while significantly reducing computational and storage costs.

Recently, diffusion models have achieved impressive results in dataset distillation by synthesizing high-quality images. In this work, we adopt Stable Diffusion \cite{rombach2022high,su2024d}, which consists of three core components: a variational autoencoder (VAE), a text encoder (CLIP), and a U-Net. The VAE comprises an encoder \(E\) that maps an input image \(x\) into a latent representation \(z = E(x)\), and a decoder \(D\) that reconstructs the image as \(\hat{x} = D(z)\). The text encoder projects label text into the latent space, providing semantic guidance for image generation. 
The U-Net serves as a conditional denoising network that predicts the noise in the noisy latent variable \( z_t \), obtained by perturbing the latent representation \( z \) with Gaussian noise at timestep \( t \), conditioned on the text-derived embedding \( c \).
The training objective minimizes the discrepancy between the predicted noise \(\epsilon_\theta(z_t, c)\) and the ground-truth noise \(\epsilon\), formulated as:
\begin{equation}
\mathcal{L}_{\mathrm{LDM}} = \left\| \epsilon_\theta(z_t, c) - \epsilon \right\|_2^2.
\end{equation}

\subsection{Framework Description}

As shown in Fig. \ref{fig2}, our method consists
of two key modules: prototype-guided image synthesis and anomaly detection with iterative refinement. The first employs a diffusion model to synthesize images guided by image prototypes. A pre-trained feature extractor is first used to obtain features for each class, which are clustered using K-means; the resulting cluster centers serve as class prototypes. Subsequently, label texts are projected into the latent space and combined with the prototypes to guide image generation via the diffusion model. The second module performs anomaly detection and iterative refinement of defective synthetic images. A detector model, trained on the original dataset, is used to identify anomalies within the synthetic dataset. Specifically, samples with incorrect labels or low confidence are classified as defective. For each low-quality sample, we re-extract its prototype and corresponding label, which jointly condition the diffusion model to generate multiple candidate images.
Candidates are ranked by detector confidence scores. From the top-$k$ candidates, we select the image that is most dissimilar to the existing normal samples to promote diversity and enhance the dataset's representativeness.

\subsection{Prototype-Guided Image Synthesis}
We first employ the pre-trained encoder \(E\) to extract latent features from input images. K-means clustering is then performed to partition each class into a predefined number of clusters, determined dynamically based on the images-per-class (IPC) setting. For example, when IPC = 10, each class is divided into 10 clusters. The resulting cluster centers serve as image prototypes, providing compact yet informative representations that enhance the efficiency and quality of dataset distillation.

These image prototypes, along with their corresponding class labels, are subsequently incorporated into a latent diffusion model (LDM) to synthesize diverse and representative images. Specifically, the LDM employs a text encoder \(\tau_\theta\) to project label information \(L\) into the latent space, which conditions the U-Net to guide image generation. For each prototype, the generation process is defined as:
\begin{equation}
\text{output} = D\left(U_t\left(\text{Concat}(z_t^c, \tau_\theta(L))\right)\right),
\end{equation}
where \(D\) is the decoder, \(z_t^c\) denotes the noisy prototype, and \(L\) represents the encoded label text.

Our refinement mechanism relies on this prototype-conditioned generation process, which enables multiple candidate samples to be re-generated from the same prototype. Diffusion models such as D\(^4\)M naturally support this form of conditioning and do not require any additional finetuning for each dataset. In contrast, Minimax Diffusion generates images directly from Gaussian noise and requires time-consuming finetuning for every target dataset. Although its finetuned images generally have correct labels and high confidence, it cannot regenerate prototype-specific variants. As a result, Minimax Diffusion is not compatible with the prototype-driven refinement pipeline proposed in this work.

\subsection{Anomaly Detection in Synthetic Images}
Although diffusion-based methods can generate visually plausible samples, the resulting synthetic images may still contain mislabeled or low-confidence instances, which can compromise downstream performance. To detect such anomalies, we employ a detector model trained on the original dataset using CutMix augmentation \cite{yun2019cutmix}, where a random region of image $x_i$ is replaced by a patch from image $x_j$:
\begin{equation}
\tilde{x}_i = \text{CutMix}(x_i, x_j, \lambda), \quad \tilde{y}_i = \lambda y_i + (1 - \lambda) y_j,
\end{equation}
where $\lambda \sim \mathrm{Beta}(\alpha, \alpha)$. The model is optimized using the following loss:
\begin{equation}
\mathcal{L}_{\mathrm{mix}} = \mathrm{CE}\left( f(\tilde{x}_i), \tilde{y}_i \right),
\end{equation}
with $\mathrm{CE}(\cdot)$ denoting the cross-entropy loss.

After training, this detector is subsequently applied to the synthetic images for anomaly detection. For CIFAR-10, we adopt the publicly available pre-trained model provided by \cite{sun2024rded} to perform this anomaly detection. A synthetic sample is classified as defective if its predicted label differs from the intended class, 
or if the softmax probability assigned to the target class falls below a threshold $\beta$. This confidence-based screening strategy follows the common practice of using softmax score to select reliable target samples~\cite{zou2025incremental}. This filtering process effectively removes low-quality or unreliable samples, enhancing the semantic integrity and overall quality of the distilled dataset.

\begin{algorithm}[h!]
\begin{algorithmic}[1]
\STATE \textbf{Input:} $(R, L)$: Real images and their labels; $DM$: Diffusion model; $f$: Pre-trained classifier
\STATE \textbf{$DM$ Components:} $E$: Encoder, $D$: Decoder, $\tau_\theta$: Text encoder, $U_t$: Time-conditional U-Net
\STATE Initialize normal set $\mathcal{N} \leftarrow \emptyset$

\FOR{each class}
    \STATE Extract feature representations using the encoder $E$.
    \STATE Perform K-Means clustering to partition the class into $C$ clusters, using the centroid $z^c$ of each cluster as the image prototype.
    \FOR{each prototype $z^c$}
        \STATE $z^c_t \sim q(z_t^c \mid z^c)$ \COMMENT{Diffusion process}
        \STATE $\tilde{z}^c = U_t(\text{Concat}(z^c_t, \tau_\theta(L)))$ \COMMENT{Denoising process}
    \ENDFOR
\ENDFOR
\STATE $S = D(\{\tilde{z}^c\})$ \COMMENT{Decode latents to generate synthetic images}

\FOR{each synthetic sample $\tilde{x} \in S$ (with corresponding prototype $z^c$ and label $L$)}
    \STATE Predict label $\hat{l}$ and confidence $p = softmax(f(\tilde{x}))_{L}$
    \IF{$\hat{l} = L$ and $p > \beta$}
        \STATE Add $\tilde{x}$ to the normal set $\mathcal{N}$
    \ELSE
        \STATE Generate candidate set $\{\tilde{x}_i\}_{i=1}^{20}$ from $z^c$ and $L$
        \STATE For each $\tilde{x}_i$, predict $\hat{l}_i$ and compute confidence $p_i = softmax(f(\tilde{x}_i))_{L}$
        \STATE Define candidate index set $\mathcal{I} = \{ i \mid \hat{l}_i = L \ \text{and} \ p_i > \beta \}$
        \IF{$\mathcal{I} \neq \emptyset$}
            \STATE Select top-$k$ indices in $\mathcal{I}$ based on $p_i$ and form candidate pool $\{\tilde{x}_i\}_{i \in \mathcal{I}_k}$
        \ELSE
            \STATE $i^{\max} = \arg\max_i p_i$ 
            \STATE Candidate pool $\{\tilde{x}_i\}_{i \in \mathcal{I}_k} \leftarrow \{\tilde{x}_{i^{\max}}\}$
        \ENDIF
        \STATE Compute feature-level similarity:
        \STATE $\tilde{x}^* = \arg\min_{\tilde{x}_i \in \{\tilde{x}_i\}_{i \in \mathcal{I}_k}} \sum_{n_j \in \mathcal{N}} \cos(v_i, n_j)$, where $v_i$ is the feature of $\tilde{x}_i$
        \STATE Replace $\tilde{x}$ with $\tilde{x}^*$ and update $\mathcal{N}$
    \ENDIF
\ENDFOR

\STATE \textbf{Output:} $\mathcal{N}$: Refined distilled dataset
\end{algorithmic}
\caption{Enhancing Dataset Distillation with Anomaly Detection}
\label{alg:1}
\end{algorithm}

\subsection{Refinement of Defective Samples}
To improve the quality of defective samples, we re-extract their image prototypes and combine them with the corresponding label texts to guide the diffusion model in generating multiple candidate images. For each defective sample, we generate 20 candidates, which are subsequently ranked according to their softmax probability for the target class predicted by the detector. To ensure the accuracy of the synthesized samples, candidates are selected only if they simultaneously meet two criteria: (1) ranking within the top-$k$ in confidence, and (2) having a softmax score that exceeds the predefined threshold $\beta$. If no candidate satisfies the above conditions, we fall back to a deterministic rule and select the candidate with the highest softmax confidence. This guarantees that each prototype contributes at least one refined sample, even in challenging cases. Among these candidates, we select the image exhibiting the greatest dissimilarity in feature space compared to previously accepted normal samples of the same class. Specifically, let \(\mathbf{v}_i\) denote the feature vector of the \(i\)-th candidate image, and let \(\mathcal{N} = \{\mathbf{n}_1, \mathbf{n}_2, \dots, \mathbf{n}_M\}\) represent the set of feature vectors of all previously accepted normal samples within the same class. The cumulative cosine similarity between \(\mathbf{v}_i\) and all elements in \(\mathcal{N}\) is computed as:
\begin{equation}
\text{Sim}(\mathbf{v}_i) = \sum_{j=1}^{M} \cos(\mathbf{v}_i, \mathbf{n}_j),
\end{equation}
where \(\cos(\cdot, \cdot)\) denotes the cosine similarity between two feature vectors. The candidate with the lowest cumulative similarity is selected:
\begin{equation}
\mathbf{v}^* = \arg\min_{\mathbf{v}_i} \text{Sim}(\mathbf{v}_i),
\end{equation}
The selected image is then added to \(\mathcal{N}\), enabling dynamic refinement of the normal sample pool for each class. This iterative process enhances both the diversity and representativeness of the distilled dataset.

\section{Experiments}
\subsection{Datasets}
We evaluate the effectiveness of our proposed method on two widely used benchmarks: CIFAR-10 \cite{krizhevsky2009learning} and ImageNette \cite{howardsmaller}, and ImageWoof, all consisting of 10 classes. CIFAR-10 is a low-resolution dataset with images of size $32\times32$ pixels. ImageNette and ImageWoof are both high-resolution datasets. ImageNette contains relatively distinct classes that are easier to distinguish, whereas ImageWoof includes classes with higher visual similarity, making it a more challenging benchmark.
\subsection{Implementation Details}
All experiments are conducted on a single NVIDIA RTX A6000 GPU. For reliability, each experiment is repeated three times with different random seeds, and the average Top-1 accuracy is reported. For training models on the original datasets, we adopt the Adam optimizer with a learning rate of 0.01 for 300 epochs. Synthetic images are generated using the stable-diffusion-v1-5 model, with a guidance scale of 10 and a strength parameter of 0.7. The resolution of the generated samples is set to $256 \times 256$ pixels for ImageNette and $32 \times 32$ pixels for CIFAR-10.
\begin{table*}[t]
\centering
\caption{Comparison of state-of-the-art methods on ImageWoof under various IPC settings and model architectures. All the results are obtained at a resolution of 256 $\times$ 256. The best results are marked as bold, and the second-best are underlined.}
\setlength{\arrayrulewidth}{0.4pt}

\resizebox{\textwidth}{!}{%
\begin{tabular}{ccccccccccccc}
\toprule
\textbf{IPC (Ratio)} & \textbf{Test Model} & \textbf{Random} & \textbf{K-Center} & \textbf{Herding} & \textbf{DiT} & \textbf{DM} & \textbf{IDC-1} & \textbf{GLaD} & \textbf{Minimax} & \textbf{D\textsuperscript{4}M} & \textbf{Ours} & \textbf{Full} \\ \midrule
 & ConvNet-6 & 24.3\scriptsize{±1.1} & 19.4\scriptsize{±0.9} & 26.7\scriptsize{±0.5} & \textbf{34.2\scriptsize{±1.1}} & 26.9\scriptsize{±1.2} & 33.3\scriptsize{±1.1} & \underline{33.8\scriptsize{±0.9}} & 33.3\scriptsize{±1.7} & 28.3\scriptsize{±2.1} & 29.9\scriptsize{±0.5} & 86.4\scriptsize{±0.2} \\
 10 (0.8\%) & ResNetAP-10 & 29.4\scriptsize{±0.8} & 22.1\scriptsize{±0.1} & 32.0\scriptsize{±0.3} & 34.7\scriptsize{±0.5} & 30.3\scriptsize{±1.2} & \textbf{39.1\scriptsize{±0.5}} & 32.9\scriptsize{±0.9} & \underline{36.2\scriptsize{±3.2}} & 33.4\scriptsize{±1.5} & 33.7\scriptsize{±1.1} & 87.5\scriptsize{±0.5} \\
   & ResNet-18 & 27.7\scriptsize{±0.9} & 21.1\scriptsize{±0.4} & 30.2\scriptsize{±1.2} & 34.7\scriptsize{±0.4} & 33.4\scriptsize{±0.7} & \textbf{37.3\scriptsize{±0.2}} & 31.7\scriptsize{±0.8} & \underline{35.7\scriptsize{±1.6}} & 32.5\scriptsize{±1.5} & 33.7\scriptsize{±0.6} & 89.3\scriptsize{±1.2} \\ \midrule
 & ConvNet-6 & 29.1\scriptsize{±0.7} & 21.5\scriptsize{±0.8} & 29.5\scriptsize{±0.3} & \underline{36.1\scriptsize{±0.8}} & 29.9\scriptsize{±1.0} & 35.5\scriptsize{±0.8} & - & \textbf{37.3\scriptsize{±0.1}} & 32.6\scriptsize{±1.4} & 35.1\scriptsize{±1.4} & 86.4\scriptsize{±0.2} \\
  20 (1.6\%) & ResNetAP-10 & 32.7\scriptsize{±0.4} & 25.1\scriptsize{±0.7} & 34.9\scriptsize{±0.1} & 41.1\scriptsize{±0.8} & 35.2\scriptsize{±0.6} & \textbf{43.4\scriptsize{±0.3}} & - & \underline{43.3\scriptsize{±2.7}} & 39.8\scriptsize{±1.1} & 41.6\scriptsize{±1.9} & 87.5\scriptsize{±0.5} \\
   & ResNet-18 & 29.7\scriptsize{±0.5} & 23.6\scriptsize{±0.3} & 32.2\scriptsize{±0.6} & \underline{40.5\scriptsize{±0.5}} & 29.8\scriptsize{±1.7} & 38.6\scriptsize{±0.2} & - & \textbf{41.8\scriptsize{±1.9}} & 39.0\scriptsize{±1.6} & 40.1\scriptsize{±0.8} & 89.3\scriptsize{±1.2} \\ \midrule
 & ConvNet-6 & 41.3\scriptsize{±0.6} & 36.5\scriptsize{±1.0} & 40.3\scriptsize{±0.7} & 46.5\scriptsize{±0.8} & 44.4\scriptsize{±1.0} & 43.9\scriptsize{±1.2} & - & \underline{50.9\scriptsize{±0.8}} & 47.9\scriptsize{±1.0} & \textbf{51.1\scriptsize{±1.0}} & 86.4\scriptsize{±0.2} \\
  50 (3.8\%) & ResNetAP-10 & 47.2\scriptsize{±1.3} & 40.6\scriptsize{±0.4} & 49.1\scriptsize{±0.7} & 49.3\scriptsize{±0.2} & 47.1\scriptsize{±1.1} & 48.3\scriptsize{±1.0} & - & 53.9\scriptsize{±0.7} & \underline{54.5\scriptsize{±0.5}} & \textbf{55.3\scriptsize{±0.8}} & 87.5\scriptsize{±0.5} \\
   & ResNet-18 & 47.9\scriptsize{±1.8} & 39.6\scriptsize{±1.0} & 48.3\scriptsize{±1.2} & 50.1\scriptsize{±0.5} & 46.2\scriptsize{±0.6} & 48.3\scriptsize{±0.8} & - & 53.7\scriptsize{±0.6} & \underline{57.5\scriptsize{±2.0}} & \textbf{58.3\scriptsize{±0.4}} & 89.3\scriptsize{±1.2} \\ \midrule
 & ConvNet-6 & 46.3\scriptsize{±0.6} & 38.6\scriptsize{±0.7} & 46.2\scriptsize{±0.6} & 50.1\scriptsize{±1.2} & 47.5\scriptsize{±0.8} & 48.9\scriptsize{±0.7} & - & 51.3\scriptsize{±0.6} & \underline{52.8\scriptsize{±0.7}} & \textbf{53.4\scriptsize{±0.9}} & 86.4\scriptsize{±0.2} \\
  70 (5.4\%) & ResNetAP-10 & 50.8\scriptsize{±0.6} & 45.9\scriptsize{±1.5} & 53.4\scriptsize{±1.4} & 54.3\scriptsize{±0.9} & 51.7\scriptsize{±0.8} & 52.8\scriptsize{±1.8} & - & 57.0\scriptsize{±0.2} & \underline{57.2\scriptsize{±1.4}} & \textbf{57.9\scriptsize{±0.7}} & 87.5\scriptsize{±0.5} \\
   & ResNet-18 & 52.1\scriptsize{±1.0} & 44.6\scriptsize{±1.1} & 49.7\scriptsize{±0.8} & 51.5\scriptsize{±1.0} & 51.9\scriptsize{±0.8} & 51.1\scriptsize{±1.7} & - & 56.5\scriptsize{±0.8} & \underline{58.4\scriptsize{±1.0}} & \textbf{59.7\scriptsize{±0.6}} & 89.3\scriptsize{±1.2} \\ \midrule
 & ConvNet-6 & 52.2\scriptsize{±0.4} & 45.1\scriptsize{±0.5} & 54.4\scriptsize{±1.1} & 53.4\scriptsize{±0.3} & 55.0\scriptsize{±1.3} & 53.2\scriptsize{±0.9} & - & \underline{57.8\scriptsize{±0.9}} & 56.3\scriptsize{±1.3} & \textbf{59.8\scriptsize{±1.2}} & 86.4\scriptsize{±0.2} \\
 100 (7.7\%) & ResNetAP-10 & 59.4\scriptsize{±1.0} & 54.8\scriptsize{±0.2} & 61.7\scriptsize{±0.9} & 58.3\scriptsize{±0.8} & 56.4\scriptsize{±0.8} & 56.1\scriptsize{±0.9} & - & \underline{62.7\scriptsize{±1.4}} & 59.7\scriptsize{±0.6} & \textbf{62.9\scriptsize{±1.0}} & 87.5\scriptsize{±0.5} \\
   & ResNet-18 & 61.5\scriptsize{±1.3} & 50.4\scriptsize{±0.4} & 59.3\scriptsize{±0.7} & 58.9\scriptsize{±1.3} & 60.2\scriptsize{±1.0} & 58.3\scriptsize{±1.2} & - & \underline{62.7\scriptsize{±0.4}} & 62.3\scriptsize{±0.8} & \textbf{65.0\scriptsize{±1.6}} & 89.3\scriptsize{±1.2} \\
\bottomrule
\end{tabular}%
}

\label{tab1}
\end{table*}

For ImageNette, we employ ResNetAP-10, a 10-layer ResNet variant in which strided convolutions are replaced with average pooling for downsampling, following \cite{gu2024efficient}. For ImageWoof, we use a ResNet-18 classifier trained under the same protocol as ImageNette. For CIFAR-10, we follow the validation protocol of RDED \cite{sun2024rded}, using a modified ResNet-18 architecture. The modifications include replacing the original $7 \times 7$ convolution with a $3 \times 3$ convolution (stride 1, padding 1), and removing the max-pooling layer after the first convolution. These adjustments preserve higher spatial resolution in early layers, which is beneficial for modeling low-resolution datasets. The hyper-parameter settings used in our experiments are as follows: top-$k = 2$ and $\beta = 0.9$ for ImageNette and ImageWoof, and top-$k = 2$ and $\beta = 0.7$ for CIFAR-10.

\subsection{Results}
We compare our method against several representative dataset distillation approaches. For ImageNette, the the baselines include non-generative methods such as DM \cite{zhao2020dataset}, IDC-1 \cite{kim2022dataset}, Herding \cite{welling2009herding}, and K-Center \cite{sener2017active}, as well as generative methods, including Minimax \cite{gu2024efficient}, $\text{D}^4\text{M}$ \cite{su2024d}, GLaD \cite{cazenavette2023glad}, and DiT \cite{gu2024efficient,peebles2023scalable}. For CIFAR-10, we compare our approach against decoupled distillation baselines, including SRe$^2$L \cite{yin2023squeeze} and RDED \cite{sun2024rded}, in addition to the generative method $\text{D}^4\text{M}$ \cite{su2024d}. Random denotes a simple baseline that randomly samples IPC images per class directly from the source dataset without any distillation. We report the mean and standard deviation of Top-1 accuracy over three independent trials. All experiments are conducted under identical conditions to ensure fair comparison. Minimax results are reproduced using its publicly available code.

\begin{table}[t!]
\centering
\captionsetup{justification=justified}

\caption{Comparison of state-of-the-art methods on ImageNette under different IPC settings, with results produced on ResNetAP-10.}
\label{tab2}
\resizebox{0.48\textwidth}{!}{%
\begin{tabular}{ccccccccc}
\toprule
& \textbf{IPC} & \textbf{Random} & \textbf{DiT} & \textbf{DM} & \textbf{D\textsuperscript{4}M} & \textbf{Minimax} & \textbf{Ours} & \textbf{Full}\\ \midrule
 & 10 & 54.2\scriptsize{±1.6} & 59.1\scriptsize{±0.7} & 60.8\scriptsize{±0.6} & 59.3\scriptsize{±2.0} & 57.7\scriptsize{±1.2} & \textbf{61.7\scriptsize{±1.9}} & \multirow{3}{*}{92.6} \\
 \textbf{Nette} & 20 & 63.5\scriptsize{±0.5} & 64.8\scriptsize{±1.2} & 66.5\scriptsize{±1.1} & 68.3\scriptsize{±0.1} & 64.7\scriptsize{±0.8} & \textbf{70.6\scriptsize{±1.6}}\\
  & 50 & 76.1\scriptsize{±1.1} & 73.3\scriptsize{±0.9} & 76.2\scriptsize{±0.4} & 76.5\scriptsize{±1.6} & 73.9\scriptsize{±0.3} & \textbf{77.7\scriptsize{±1.7}}\\ \bottomrule
\end{tabular}%
}
\end{table}
\begin{table}[t!]
\centering
\caption{Comparison of performance on CIFAR-10 under different IPC settings.}
\label{tab:cifar}
\resizebox{0.48\textwidth}{!}{
\begin{tabular}{ccccccc}
\toprule
\textbf{Dataset} & \textbf{IPC} & \textbf{SRe\textsuperscript{2}L} & \textbf{RDED} & \textbf{D\textsuperscript{4}M} & \textbf{Ours} & \textbf{Full} \\
\midrule
\multirow{2}{*}{CIFAR-10} 
 & 10 & 29.3\scriptsize{±0.5} & 37.1\scriptsize{±0.3} & 36.1\scriptsize{±1.5} & \textbf{39.8\scriptsize{±0.9}} & \multirow{2}{*}{91.4} \\
 & 50 & 45.0\scriptsize{±0.7} & 62.1\scriptsize{±0.1} & 66.0\scriptsize{±1.0} & \textbf{66.5\scriptsize{±1.6}}\\
\bottomrule
\end{tabular}
}
\end{table}
\textbf{ImageWoof} We evaluate our method under various IPC settings using three architectures: ConvNet-6, ResNetAP-10, and ResNet-18, as reported in Tab.~\ref{tab1}. Across all settings and models, our approach consistently outperforms the D\textsuperscript{4}M baseline, achieving an average improvement of 1.7\% in top-1 accuracy. Under the extremely low data regime (IPC = 10), our method outperforms D\textsuperscript{4}M by 1.0\% in terms of average accuracy. As the IPC increases, the performance gap becomes more pronounced. At IPC = 50 and IPC = 70, our approach yields average gains of 1.6\% and 1.9\% over D\textsuperscript{4}M, respectively. Notably, under the high IPC setting (IPC = 100), our method achieves an average gain of 3.1\% over D\textsuperscript{4}M across all architectures. In particular, it attains 65.0\% accuracy with ResNet-18, exceeding the performance of D\textsuperscript{4}M by 3.5\%. Compared with the state of the art Minimax baseline, our method is less competitive under low IPC settings. However, it begins to outperform Minimax once more distilled samples become available. At IPC = 50, IPC = 70, and IPC = 100, our method surpasses Minimax by 2.1\%, 2.1\%, and 1.5\% on average across the three architectures. The gains are particularly pronounced for ResNet-18, with improvements of 4.6\% at IPC = 50, 3.2\% at IPC = 70, and 2.3\% at IPC = 100. These results demonstrate that our method gains performance more effectively with increasing IPC and achieves performance competitive with existing state-of-the-art baselines.

\textbf{ImageNette} As shown in Tab.~\ref{tab2}, we evaluate our method under varying Images Per Class (IPC) settings of 10, 20, and 50. Across all IPC settings, our method consistently outperforms $\text{D}^4\text{M}$ and other competitive baselines. Specifically, we observe relative improvements of 2.4\%, 2.3\%, and 1.2\% over the baseline $\text{D}^4\text{M}$ at IPC = 10, 20, and 50, respectively. These results demonstrate that our anomaly-aware refinement effectively mitigates the negative impact of defective synthetic samples, thereby improving downstream classification performance.
It is worth noting that the performance gain of our method over $\text{D}^4\text{M}$ decreases as the IPC increases. For example, while we achieve a substantial improvement of 2.4\% at IPC = 10, the relative advantage decreases to 1.2\% at IPC = 50. This diminishing margin is expected, as larger IPC settings naturally yield more diverse and informative training data, which reduces the adverse impact of low-quality samples even for baseline methods. In contrast, under low IPC settings, the total number of synthetic samples is inherently limited, meaning that even a small number of low-quality or mislabeled instances can account for a large proportion of the dataset. Consequently, identifying and rectifying such defective samples exerts a disproportionately large impact on model generalization.

\begin{figure}[t] 
\centering 
\captionsetup{justification=raggedright}
\includegraphics[width=0.45\textwidth]{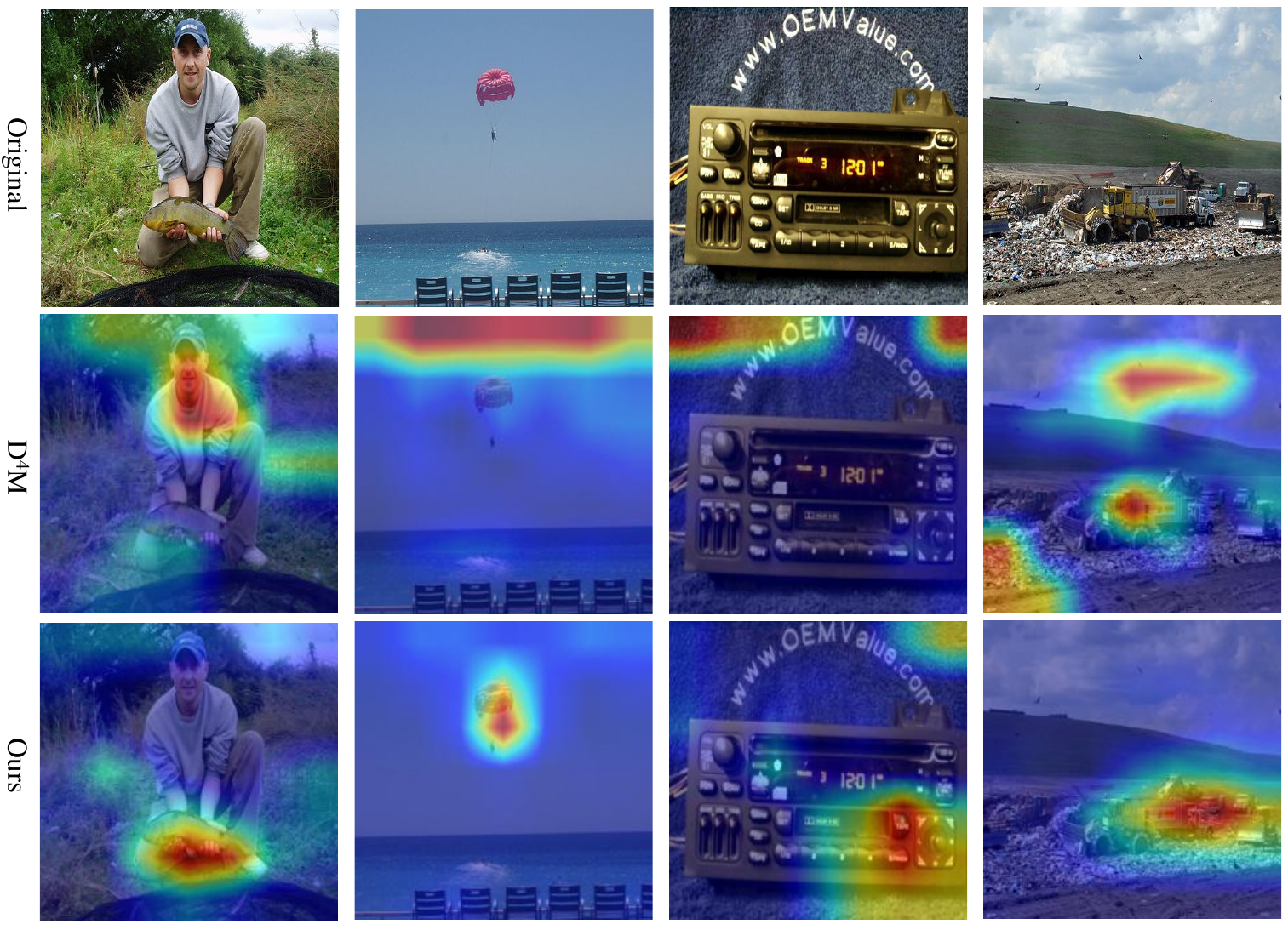} 
\caption{
Comparison of Grad-CAM visualizations across different methods. The first row shows the original input images. The second row presents attention maps generated by D\textsuperscript{4}M, while the third row shows results from our proposed method.
}
\label{fig4} 
\end{figure}
\textbf{CIFAR-10} We further evaluate our method on the low-resolution CIFAR-10 dataset under IPC settings of 10 and 50, as shown in Tab.~\ref{tab:cifar}. In both scenarios, our method consistently surpasses existing baselines, including SRe$^2$L~\cite{yin2023squeeze}, RDED~\cite{sun2024rded}, and $\text{D}^4\text{M}$~\cite{su2024d}. Specifically, our method achieves 39.8\% Top-1 accuracy at IPC = 10, outperforming $\text{D}^4\text{M}$ by 3.7\% and RDED by 2.7\%. At IPC = 50, we obtain 66.5\%, exceeding $\text{D}^4\text{M}$ by 0.5\%. 
Consistent with the observations on ImageNette, the relative performance gains decrease as IPC increases. Nevertheless, despite this diminishing margin, our method consistently maintains a performance advantage across all IPC settings, with particularly notable improvements under low IPC conditions, where defective samples exert a greater influence on model performance. These results further validate the effectiveness and generalizability of our anomaly-aware refinement strategy across both high- and low-resolution datasets.
\begin{figure*}[t] 
\centering 
\captionsetup{justification=raggedright}
\includegraphics[width=0.85\textwidth]{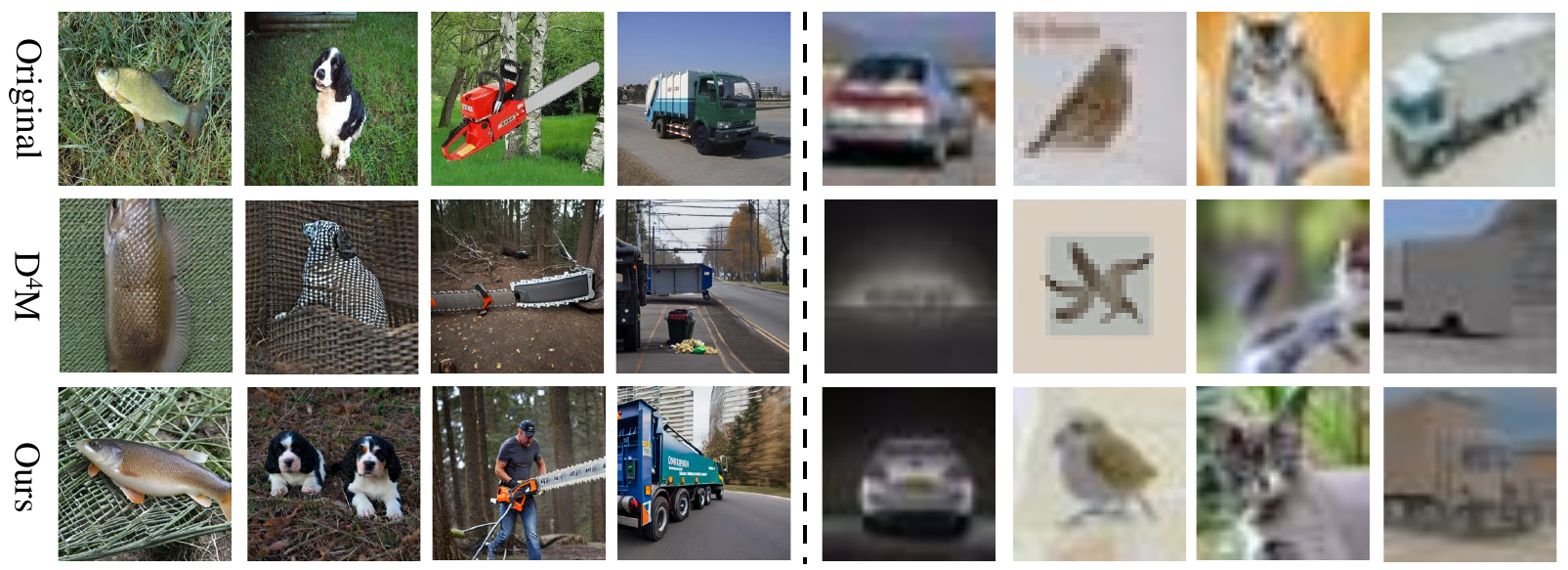} 
\caption{
Visualization of images generated using different semantic strategies. For each column, the images are generated using the same
image prototype and random seed. The left panel displays results on ImageNette ($256 \times 256$ pixels), and the right panel shows results on CIFAR-10 ($32 \times 32$ pixels). Compared to the baseline, our approach generates images with more complete semantic structures and clearer class-discriminative features.
}
\label{fig3} 
\end{figure*}
\begin{figure*}[!t]
\centering
\subfloat[] {\includegraphics[width=0.45\linewidth,scale=0.5]{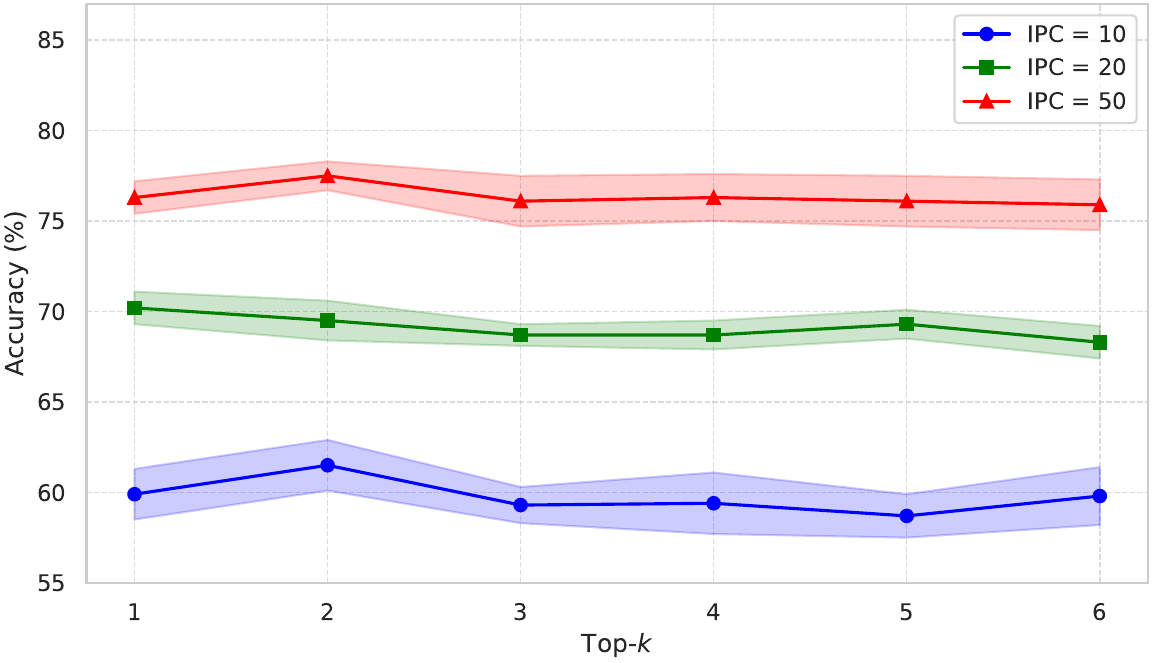}\label{fig:5a}}\hspace{0.5em}%
\subfloat[]
{\includegraphics[width=0.45\linewidth,scale=0.5]{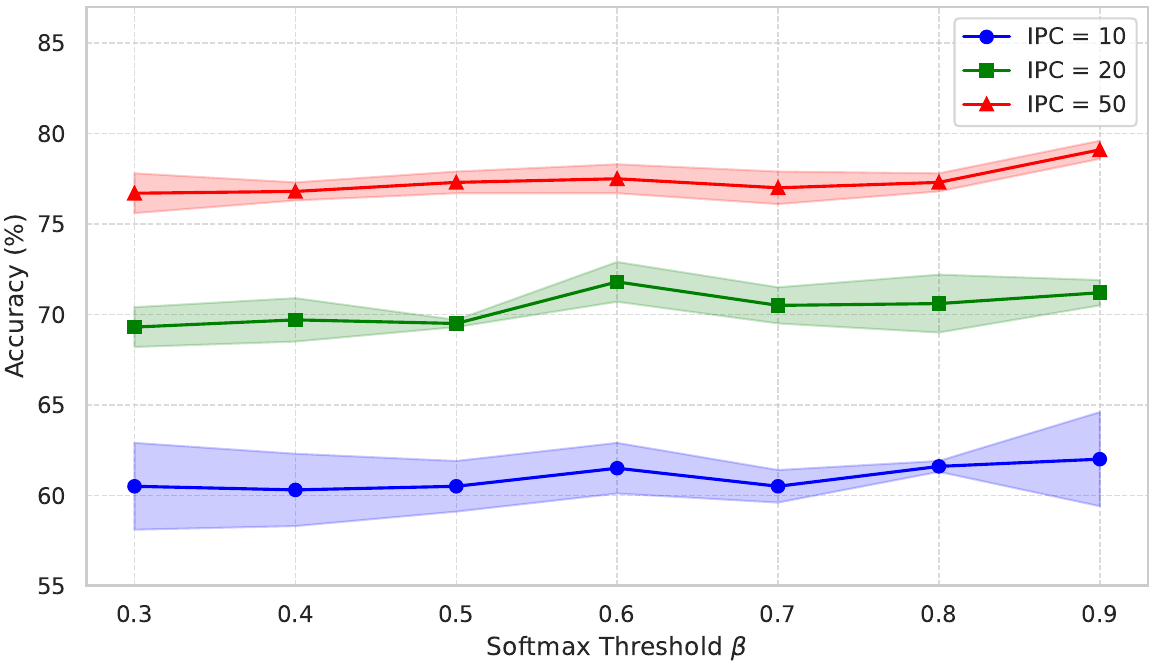}\label{fig:5b}}\hspace{0.5em}%
\caption{Parameter Analysis. (a) Effect of the top-$k$ candidate pool size on accuracy. 
(b) Effect of the softmax confidence threshold $\beta$ on accuracy.}
\label{fig:5}
\end{figure*}
\subsection{Visualization}
\subsubsection{Qualitative Comparison of Generated Samples}

We conduct qualitative comparisons to visually assess the synthesized datasets, as shown in Fig. \ref{fig3}. Each column shows samples generated from the same image prototype and random seed, allowing direct comparisons across different methods. The left panel shows results on the high-resolution ImageNette dataset ($256 \times 256$ pixels), while the right panel displays results on the low-resolution CIFAR-10 dataset ($32 \times 32$ pixels).
Compared to the baseline method $\text{D}^4\text{M}$, which often produces artifacts, structural distortions, or incomplete object representations, our method synthesizes images with structurally complete objects and clear class-discriminative details. These improvements are consistently observed across diverse categories and contribute to enhancing feature learning during downstream tasks.

\subsubsection{Grad-CAM Analysis of Model Attention}
To further evaluate the quality of the distilled datasets, we generate Grad-CAM attention maps on validation samples using models trained with different distillation methods. As shown in Fig. \ref{fig4}, a model trained on a dataset distilled by $\text{D}^4\text{M}$ often exhibits attention misalignment, where attention is erroneously assigned to background regions or non-discriminative areas. For instance, in the first column, attention is misallocated to the person rather than the intended target, a fish; similarly, in the second column, attention is incorrectly assigned to the sky instead of the parachute.  In contrast, our method generates attention maps that are accurately aligned with the target objects, resulting in improved localization of class-discriminative regions. These results indicate that our approach preserves meaningful structural information and enhances both interpretability and classification performance on downstream tasks.
\begin{table}[t]
\caption{Generation quality on ImageNette and ImageWoof at IPC = 50.}
\centering
\resizebox{0.48\textwidth}{!}{
\begin{tabular}{l l c c c c c}
\toprule
\textbf{Dataset} & \textbf{Method} & \textbf{FID $\downarrow$} & \textbf{Precision (\%) $\uparrow$} &  \textbf{Density $\uparrow$} & \textbf{Coverage (\%) $\uparrow$} \\
\midrule
\multirow{2}{*}{ImageNette} 
  & D$^{4}$M & 54.38{\scriptsize$\pm$0.94} & 75.60{\scriptsize$\pm$1.93} &  0.73{\scriptsize$\pm$0.04} & 10.30{\scriptsize$\pm$0.67} \\
  & Ours     & 52.96{\scriptsize$\pm$0.39} & 79.47{\scriptsize$\pm$0.94} &  0.79{\scriptsize$\pm$0.02} & 10.89{\scriptsize$\pm$0.51} \\
\midrule
\multirow{2}{*}{ImageWoof} 
  & D$^{4}$M & 51.60{\scriptsize$\pm$0.53} & 79.26{\scriptsize$\pm$0.75} &  0.73{\scriptsize$\pm$0.02} & 10.36{\scriptsize$\pm$0.38} \\
  & Ours     & 50.18{\scriptsize$\pm$0.95} & 81.27{\scriptsize$\pm$0.83} &  0.77{\scriptsize$\pm$0.03} & 10.74{\scriptsize$\pm$0.45} \\
\bottomrule
\end{tabular}
}
\label{tab:mgd_d4m_ours_metrics}
\end{table}
\subsection{Quantitative Evaluation of Synthetic Data}
We further assess generation quality using the metrics reported in Tab.~\ref{tab:mgd_d4m_ours_metrics}. As shown in the table, our method consistently improves the quality of the synthesized data on both ImageNette and ImageWoof compared with D\textsuperscript{4}M. On ImageNette, it reduces the FID from 54.38 to 52.96 and increases both density and coverage from 0.73 to 0.79 and from 10.30\% to 10.89\%, respectively, while also achieving a higher precision score of 79.47\% versus 75.60\% for D\textsuperscript{4}M. A similar trend is observed on ImageWoof, where our approach attains a lower FID of 50.18 compared with 51.60 for D\textsuperscript{4}M, together with improvements in precision, density, and coverage. These results indicate that our framework yields higher quality synthetic data whose distribution is more closely aligned with that of the real data on both datasets.

Beyond distribution-level quality, we also evaluate the label consistency of the synthesized samples. Specifically, on the ImageWoof dataset under the baseline setting, 10.2\% of the generated samples have incorrect labels and 5\% have confidence scores below 0.7. In contrast, with our method, only 0.2\% of the labels are incorrect, and none of the samples exhibit confidence scores below 0.7. These results clearly indicate that our refinement strategy substantially improves both label correctness and confidence reliability.
\subsection{Ablation and Sensitivity Analysis}
\begin{table}[t]
\caption{Ablation results under different IPC settings on ImageNette using ResNetAP10.}
\label{tab:ablation}
\centering
\footnotesize
\begin{tabular*}{0.48\textwidth}{@{\extracolsep{\fill}}ccccc}
\toprule
\textbf{IPC} & \textbf{Baseline} & \textbf{Conf} & \textbf{Sim} & \textbf{Conf+Sim} \\ \midrule
10 & 59.3\scriptsize{±2.0} & 59.9\scriptsize{±1.4} & 59.3\scriptsize{±1.9} & \textbf{61.7\scriptsize{±1.9}} \\
20 & 68.3\scriptsize{±0.1} & 70.2\scriptsize{±0.9} & 69.2\scriptsize{±0.8} & \textbf{70.6\scriptsize{±1.6}} \\
50 & 76.5\scriptsize{±1.6} & 76.3\scriptsize{±0.9} & 76.0\scriptsize{±0.5} & \textbf{77.7\scriptsize{±1.7}} \\
\bottomrule
\end{tabular*}%
\end{table}
\subsubsection{Ablation Study}

We conduct an ablation study in Tab.~\ref{tab:ablation} to assess the contribution of the confidence-based and similarity-based selection components in the refinement of defective samples stage. Here, Baseline denotes the original D\textsuperscript{4}M pipeline without anomaly-aware refinement. Conf is a confidence only strategy that selects candidates solely by the softmax score, while Sim is a similarity only strategy that ignores confidence and selects samples with minimum feature similarity to the normal set. Conf+Sim first forms a top$k$ candidate pool based on confidence and then selects the least similar sample within this pool. Compared to the baseline D\textsuperscript{4}M, all refined strategies yield improved accuracy in the low-IPC regime, except for Sim. Notably,  the Conf strategy consistently outperforms the baseline at both IPC = 10 and IPC = 20, achieving an improvement of 1.9\% at IPC = 20. This suggests that selecting samples solely on softmax confidence is an effective approach for identifying representative instances when training data is limited.
In contrast, the Sim strategy yields gains only at IPC = 20 but slightly underperforms at IPC = 50. This implies that relying exclusively on feature dissimilarity may be inadequate when the selected samples lack sufficient discriminative features. Among all strategies, Conf + Sim achieves the best overall performance, particularly in a data-scarce scenario. At IPC = 10, it attains 61.7\%, surpassing all other methods by a clear margin. These results highlight that jointly leveraging softmax confidence and feature-level dissimilarity enhances both the diversity and the discriminative details present in the selected samples.

\subsubsection{Parameter Analysis}
A re-generated sample is regarded as valid only if it simultaneously satisfies two criteria: 
it must rank within the top-$k$ candidates in terms of confidence among all samples drawn from the same prototype, and its softmax score must exceed a predefined threshold $\beta$. 
The former controls the size of the candidate pool, while the latter imposes a minimum confidence level that discards low-probability predictions. We conduct a sensitivity analysis on the top-$k$ candidate pool size and the softmax threshold $\beta$, as illustrated in Fig. \ref{fig:5}. For parameter top-$k$ shown in Fig.~\ref{fig:5}\subref{fig:5a}, performance peaks at $k=2$ for both IPC = 10 and IPC = 50, indicating that a small yet confident candidate pool is sufficient to preserve discriminability and promote diversity. As $k$ increases beyond this point, accuracy slightly declines, likely due to the inclusion of less informative or noisy candidates. As shown in Fig.~\ref{fig:5}\subref{fig:5b}, we also analyze the impact of the softmax confidence threshold $\beta$ on model performance. Across all IPC levels, increasing $\beta$ generally leads to improved accuracy, with the highest performance observed at $\beta = 0.9$ for IPC = 50. This trend suggests that imposing stricter confidence constraints eliminates ambiguous candidates and ensures that only highly confident predictions are retained. Consequently, the quality of the top-$k$ candidate pool is significantly improved, facilitating more reliable selection of representative samples. Overall, the results indicate that these two parameters play complementary roles in balancing confidence and diversity. A small top-$k$ value limits the candidate pool, potentially improving sample quality. However, the top-$k$ candidates may still include low-confidence predictions. Introducing a high softmax threshold $\beta$ serves as a lower bound to filter out semantically uncertain samples, thereby enhancing the reliability of the top-$k$ selection process.

\section{Conclusion}
In this work, we propose a detector-guided
dataset distillation framework that integrates diffusion-based image generation with anomaly detection to address the quality limitations of existing synthetic datasets. By introducing a detector model trained on the original dataset, our method identifies defective synthetic samples exhibiting label inconsistencies or structural flaws, which are then refined through a targeted regeneration process. Specifically, for each anomalous image, multiple variants are synthesized based on the corresponding prototype and label, and the optimal candidate is selected by jointly considering the detector's confidence score and dissimilarity to existing qualified samples. This strategy effectively improves the overall quality, label consistency, and intra-class diversity of the distilled dataset. Extensive experiments on ImageNet subsets demonstrate that the proposed method generates high-quality, structurally coherent synthetic images, achieving superior downstream performance. Nevertheless, as a limitation, the image prototypes generated by K-means exhibit limited representativeness of the original data. In future work, we will investigate more advanced prototype construction techniques to further improve the representativeness and diversity of the generated datasets.


\bibliographystyle{unsrt}
\bibliography{reference}
\end{document}